\documentclass[11pt]{article}

\usepackage[final]{acl}

\usepackage{times}
\usepackage{latexsym}

\usepackage[T1]{fontenc}

\usepackage[utf8]{inputenc}

\usepackage{microtype}

\usepackage{inconsolata}

\usepackage{graphicx}

\makeatletter
\renewcommand\@makefnmark{}   
\makeatother

\usepackage{multirow}  
\usepackage{graphicx}
\usepackage{booktabs}   
\usepackage{tabularx}   
\usepackage{adjustbox}  
\usepackage{pifont}     
\usepackage{wrapfig}  
\usepackage{algorithm}      
\usepackage{algpseudocode}  
\usepackage{caption}        
\usepackage{enumitem}
\usepackage{amsmath}
\usepackage{amssymb}

\usepackage{fontawesome5}
\usepackage{xcolor}
\usepackage{hyperref}
\newcommand{\equalcontrib}{\thanks{$^{\dagger}$ These authors contributed equally to this work.}}
\newcommand{\corresponding}{\thanks{$^{*}$ Corresponding author.}}

\hypersetup{
  colorlinks=true,
  linkcolor=black,
  citecolor=black,
  urlcolor=blue
}
\usepackage{tcolorbox}   
\usepackage{fontawesome5}   
\usepackage{xcolor}         
\definecolor{linkblue}{RGB}{0,102,204} 
\title{Helios: A Foundational Language Model for Smart Energy Knowledge Reasoning and Application}

\author{
Haoyu Jiang$^{1,\dagger}$\equalcontrib \;
Fanjie Zeng$^{2,\dagger}$ \;
Boan Qu$^{2,\dagger}$ \;
Xiaojie Lin$^{1,*}$\corresponding \;
Wei Zhong$^{1,3}$ \\
\\
$^{1}$ College of Energy Engineering, Zhejiang University \\
$^{2}$ Polytechnic Institute, Zhejiang University \\
$^{3}$Shanghai Institute for Advanced Study, Zhejiang University\\
\\
{\small
\faEnvelope[regular]\:
\href{mailto:haoyu.jiang@zju.edu.cn}{
\{haoyu.jiang, zengfanjie, boan.qu, xiaojie.lin, wzhong\}@zju.edu.cn}
\quad
\faGlobe\:
\href{https://helios-llm.github.io/}{
https://helios-llm.github.io/}
}
}

\begin{document}
\maketitle
\begin{abstract}

In the pursuit of carbon neutrality, smart energy systems integrating renewable energy, storage, and demand response have become central to energy system transformation. However, fragmented and rapidly evolving interdisciplinary knowledge increases the cognitive and information integration burden for operational decision-making. Although large language models (LLMs) have shown promise in smart energy tasks through fine-tuning or prompt engineering, their lack of domain knowledge and physical constraints often results in semantically plausible but physically inconsistent outputs, limiting their engineering reliability.
To address these challenges, we introduce \textbf{Helios}, the first large language model tailored to the smart energy domain, together with a comprehensive suite of resources to advance LLM research in this field. Specifically, we develop \textbf{EnerSys}, a multi-agent collaborative framework for end-to-end dataset construction, through which we produce: (1) the first smart energy knowledge base, \textbf{EnerBase}, to enrich the model’s foundational expertise; (2) the first instruction tuning dataset, \textbf{EnerInstruct}, to strengthen performance on domain-specific downstream tasks; and (3) the first Reinforcement Learning from Human Feedback (RLHF) dataset, \textbf{EnerReinforce}, to align the model with human preferences and industry standards. Leveraging these resources, Helios undergoes large-scale pretraining, instruction tuning, and RLHF. We also release \textbf{EnerBench}, the first benchmark for evaluating LLMs in smart energy scenarios, and demonstrate that our approach significantly enhances domain knowledge mastery, task execution accuracy, and alignment with human preferences. 

\end{abstract}

\section{Introduction}

\begin{figure}[!h]
  \centering
   \includegraphics[width=0.9\linewidth]{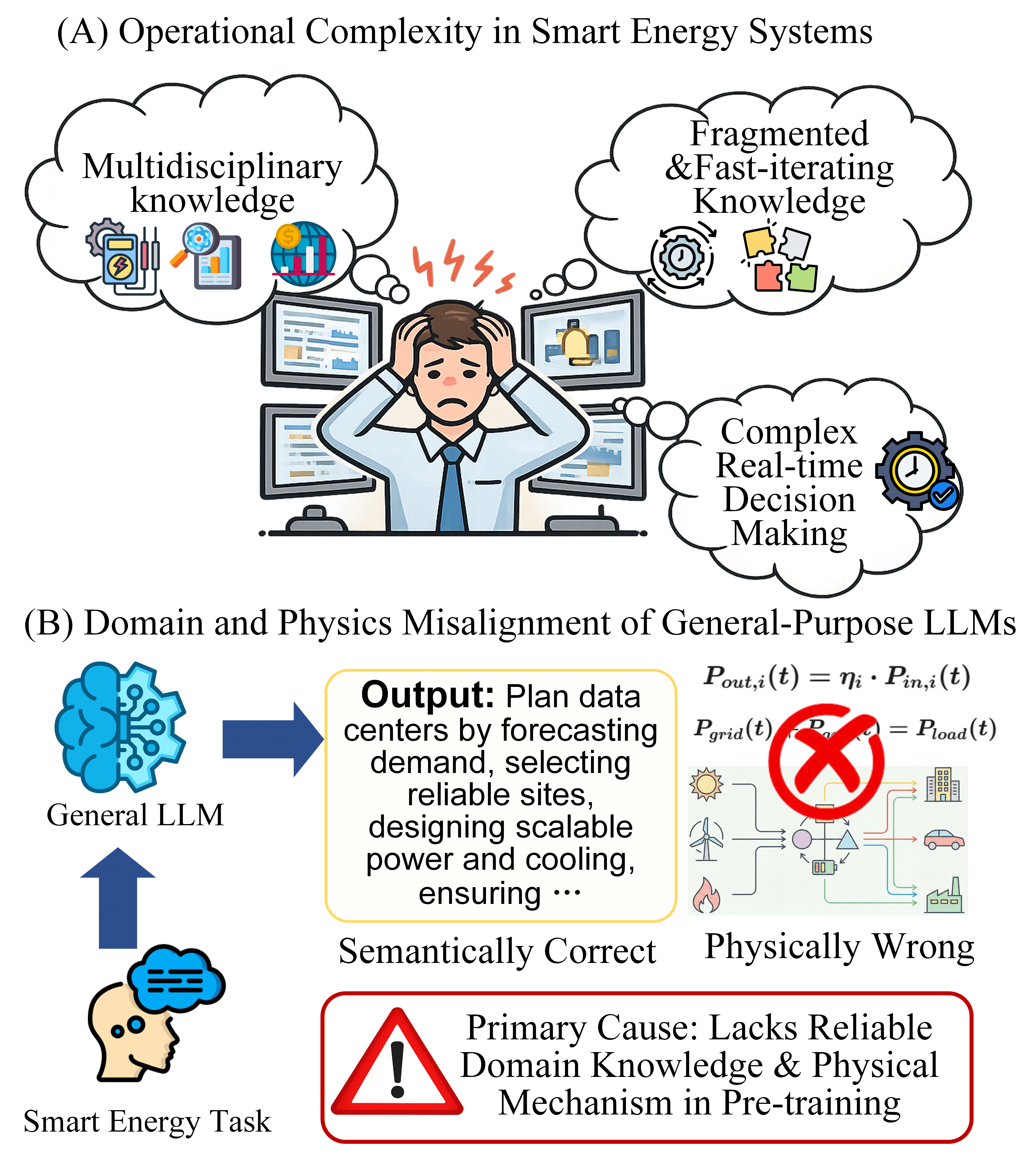}
   \vspace{-2mm}
   \caption{ Limitations of Scheduling Decision-Making in the Smart Energy Domain.} 
   \vspace{-4mm}
   \label{fig:int}
\end{figure}

Driven by the global pursuit of carbon neutrality, smart energy systems must enhance overall efficiency through the intelligent coordination of renewable energy integration, energy storage dispatch, and demand‑side response \cite{lund2017smart}. Smart energy is highly interdisciplinary, encompassing power engineering, information science, economics, and other fields, and its knowledge base is fragmented and rapidly evolving \cite{ceglia2020smart,thellufsen2020smart}. Building on recent advances in general large language models (LLMs) in semantic understanding, logical reasoning, and multitask generalization, a growing body of research has used fine‑tuning and prompt engineering to adapt LLMs to task‑specific applications in smart energy, such as load forecasting \cite{jin2023time, liao2025timegpt, hu2025applying}, building energy consumption modeling \cite{wang2025climate,jiang2024eplus}, and HVAC fault diagnosis \cite{zhang2025domain}, thereby supporting case modeling and intelligent decision making.

However, general LLMs often deliver reasoning that is semantically plausible yet physically invalid \cite{friel2023chainpoll}. This limitation arises chiefly because their pre‑training corpora lack reliable knowledge from the smart energy domain, leaving the models without essential domain context and physical constraints \cite{10.1145/3616855.3635772}. Current approaches \cite{hu2025applying, zhang2025domain} mainly invoke the prior knowledge already embedded in LLMs and do not explicitly enrich them with smart energy expertise.
To alleviate these challenges, we introduce the first-ever open‑sourced foundational LLM for the smart‑energy domain, referred to as \textbf{Helios} (Originating from the ancient Greek sun‑god, signifies the illumination of the pathway toward sustainable development through the radiance of smart energy, thereby advancing the harmonious co‑existence of humanity and the natural environment). Helios is capable of effectively tackling a broad spectrum of smart‑energy tasks.
Furthermore, we present \textbf{EnerSys}, an end-to-end multiagent collaborative framework for dataset construction that integrates automated data generation, screening, and refinement, thereby furnishing Helios with an extensive and high‑quality data foundation.

EnerSys covers three dataset‑construction phases (as shown in Fig. \ref{fig:int}): In the construction of the pre-training dataset, the Parsing‑Agent and Deduplication Agent extract structured knowledge from the Smart Energy Corpus (scientific papers, domain‑modeling code, IEA datasets, etc.) and eliminate redundancy, building a comprehensive, balanced smart energy domain knowledge base, \textbf{EnerBase}; In instruction‑tuning dataset construction, on expert‑crafted seed data, we deploy Expert‑Agents for each of 14 smart‑energy sub‑domains, letting them generate instruction–response pairs from the seeds and a high‑quality corpus; the Check‑Agent then scores samples on accuracy, completeness, relevance, and usability, and the Refine‑Agent automatically fixes those below par. This pipeline yielded the \textbf{EnerInstruct}; In the RLHF dataset construction, agents like Write‑like‑Human craft multi‑level candidate answers to given questions, thereby creating the \textbf{EnerReinforce} to supply the reward model with differentiated contrastive samples. Using these datasets, we complete Helios pre‑training (adding domain basics), supervised fine‑tuning (boosting downstream skills), and RLHF reinforcement (aligning with human preferences). Concurrently, adhering to a dual‑track paradigm of “public item‑bank retrieval + expert‑targeted design,” we build \textbf{EnerBench}, containing 625 subjective and 887 objective questions, to systematically assess LLMs performance in smart‑energy scenarios. Experiments show Helios surpasses general‑purpose LLMs on both tasks, with output style tightly matching professional context. Our contributions can be summarized as follows:

1) We design Helios, the first foundation large language model in the smart‑energy domain; it effectively tackles a wide range of smart‑energy tasks and produces outputs that are deeply consistent with professional discourse.

2) We propose EnerSys, an end‑to‑end, multi‑agent collaborative framework for dataset construction, through which we develop a domain knowledge base, an instruction‑tuning dataset, and an RLHF database for smart energy. In addition, we release Smart Energy Bench, a benchmark that systematically evaluates LLMs’ comprehensive performance in smart‑energy scenarios.

3) Relative to general LLMs, Helios delivers superior results on subjective (multiple‑choice, cloze, and judgment) and objective (essay writing, term explanation, and modelling‑and‑optimization) tasks in the smart‑energy field. 

\begin{figure*}[!h]
  \centering
   \includegraphics[width=0.97\linewidth]{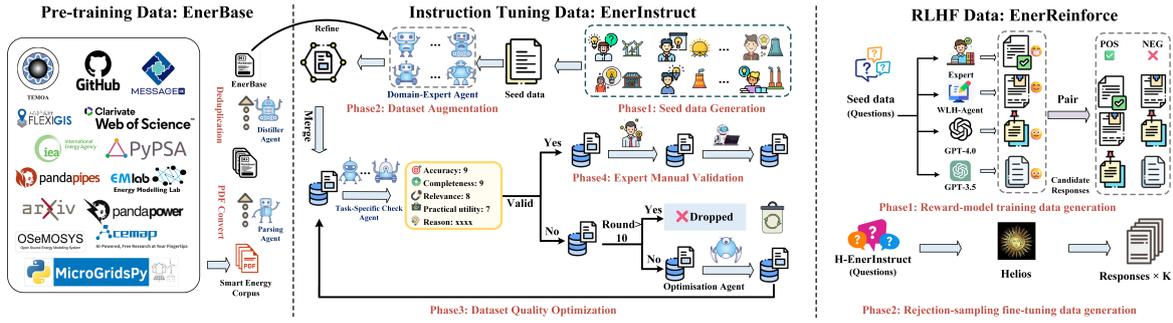}
   \vspace{-2mm}
   \caption{ The multi-agent collaboration framework \textbf{EnerSys} provides the data required for \textbf{Helios}' three-stage training, including pre-training data (EnerBase), instruction tuning data (EnerInstruct), and RLHF data (EnerReinforce).} 
   \vspace{-2mm}
   \label{fig:model}
\end{figure*}

\section{Related Work}

\textbf{Foundation Language Models.}
LLMs trained on vast amounts of diverse and heterogeneous data, have accumulated extensive domain knowledge and contextual modeling capabilities. They have demonstrated human-level performance in many tasks. LLMs can be categorized into two types: 1) Closed-source models (such as OpenAI o1 \cite{jaech2024openai} and Claude): These models provide inference interfaces via APIs, making them suitable for industrial-grade deployment without the need for building custom computational resources. However, they cannot be customized or extended according to specific needs; 2) Open-source models (such as DeepSeek \cite{guo2025deepseek, liu2024deepseek}, Llama and Qwen \cite{bai2023qwen, bai2025qwen2}): These models offer complete training weights, allowing for customized applications based on downstream task requirements. This has led to the development of instruction-tuned models like Alpaca \cite{taori2023alpaca}, Vicuna \cite{chiang2023vicuna}, and Dolly \cite{conover2023hello}. In this process, the quality of datasets becomes a critical factor affecting training outcomes.

\textbf{Domain Language Models. }LLMs excel in general reasoning \cite{jiang2025ucdr, nam2024using}, their performance in specialized applications is hampered by a lack of domain expertise. This limitation has led researchers to adapt foundation models for vertical domains such as medicine \cite{tian-etal-2024-chimed,Lin2025HealthGPTAM}, chemistry \cite{zhang2024chemllmchemicallargelanguage,zheng2025large}, ocean science \cite{bi-etal-2024-oceangpt}, and geography \cite{10.1145/3616855.3635772}. However, most domains are still in the early stages of exploration. In the energy sector, existing research primarily leverages general models' prior knowledge through prompt engineering or fine-tuning for load forecasting \cite{jin2023time, jiang2025hyperloadcrossmodalityenhancedlarge, wu2024stellm, wang2025chattime}, building energy modeling \cite{wang2025climate,jiang2024eplus}, teaching assistant \cite{lin2025generative}, and HVAC fault diagnosis \cite{zhang2025domain}. These approaches focus on application (applying LLMs' prior knowledge to downstream tasks) rather than accumulation (enriching models with energy domain knowledge through pretraining). Furthermore, the energy domain faces a scarcity of high-quality training data due to literature repository access restrictions and computational resource costs \cite{wang2022few, chen2024novel}. The current instruction fine-tuning data construction method \cite{wang2023self, zhang2023instruction}, which relies heavily on large model generation, amplifies discrepancies between response styles and human preferences, posing a significant challenge. This paper introduces the first energy domain-specific large language model, a novel development that completes full-process training, constructs the domain's first training and evaluation datasets, and enhances model response alignment with human preferences through RLHF.

\textbf{Multi-Agent Systems.}
 Due to the extensive domain knowledge and robust semantic understanding capabilities of LLMs, they are employed as core components of agents to support intelligent decision-making and natural language interaction \cite{guo2024large}. In single-agent systems, a single agent carries out decision-making and task execution, which is suitable for structured scenarios with fewer variables \cite{chen2023agentverse}. However, as the complexity of problems increases, single-agent systems face issues such as low decision-making efficiency, slow response times, and poor fault tolerance \cite{amirkhani2022consensus}. In contrast, multi-agent systems can effectively address these challenges through the collaboration of specialized agents and have been widely applied in complex scenarios such as interactive games \cite{mao2023alympics, xu2023language}, financial markets \cite{li2023camel}, and social simulations \cite{park2023generative}. Currently, some scholars \cite{bi-etal-2024-oceangpt, ni2024mechagents} are exploring ways to improve the efficiency and representativeness of domain dataset construction through multi-agent collaboration and distributed decision-making mechanisms. However, constructing domain datasets typically involves multiple steps, including data generation, deduplication, filtering, and optimization. Existing research often focuses only on optimizing specific steps and has not fully leveraged the potential of multi-agent systems in dataset construction.

\section{Data Collection and Curation}
To meet the stringent high-quality data requirements of Helios during the pre-training, instruction tuning, and RLHF stages, we have designed an efficient multi-agent collaborative dataset construction framework, EnerSys (see Fig. \ref{fig:model}).

\subsection{Pre-training Data: EnerBase}
In this work, we conducted specialized text data pre-training based on the Qwen2.5-7B foundation model. The constructed Smart Energy Corpus includes open-access academic preprints, authoritative journal papers, specialized publications, domain-specific modeling toolkits, and application code, and IEA energy datasets from the smart energy domain. Data was collected from arXiv, Web of Science (WoS), Acemap, Github, and HuggingFace platforms. After data preprocessing, we constructed \textbf{EnerBase}, a high-quality pre-training corpus of \textit{approximately 3 billion tokens} to enhance the model's accumulation of professional knowledge and technical application capabilities in the smart energy domain.
In brief, the statistical characteristics of Smart Energy Corpus are shown in Table \ref{tab:corpus_stats}.

\subsubsection{Smart Energy Corpus Collection}
\textbf{Scientific Literature.}
The smart energy domain's extensive scientific literature provides a high-quality training corpus for LLMs, enhancing their domain-specific knowledge understanding and application capabilities. To ensure the comprehensiveness of the corpus, we systematically decomposed the smart energy domain into 14 specialized sub-domains, and collected data for each separately.
\textbf{1) Open-access Academic Preprints (OAP):}
We crawled 173,541 PDF files from arXiv using subdomain-specific keywords, establishing the quantitative foundation of our Smart Energy Corpus.
\textbf{2) Open-access Authoritative Journal Papers (OAJP):}
We extracted metadata from WoS for leading energy journals and crawled 32,459 PDF files, establishing the \emph{qualitative foundation} of our Smart Energy Corpus.
\textbf{3) Specialized Publications (SP):}
We crawled 363 book PDF files from the Acemap, enriching the Smart Energy Corpus knowledge framework.

\textbf{Domain-specific Modeling Toolkits and Application Code (DMT\&AC).}
Modern smart energy systems face exponential complexity due to multi-dimensional coupling of renewable integration, demand-side response, and power-carbon market mechanisms. Researchers employ high-precision algorithms and parallel computing for large-scale system optimization. Python dominates energy system modeling with its scientific computing ecosystem and machine learning capabilities, with 89\% of modeling tools now open-source through community development \cite{RePEc:eee:rensus:v:212:y:2025:i:c:s1364032125000401}.
To enhance language models' capabilities in parsing and generating specialized code for smart energy applications, we selected 19 representative frameworks (including Oemof \cite{Hilpert_2018}, OSeMOSYS \cite{article}, TEMOA \cite{Lerede2024TEMOAEuropeAO}) and application libraries, extracting 5,389 Python files and 278 Jupyter notebooks. 

\textbf{International Energy Agency Datasets (IEAD).}
The IEA, covering 75\% of global energy demand, has evolved from an oil crisis response mechanism to a platform governing energy security, economic growth, and environmental protection. Its statistics system provides authoritative data on supply-demand balance, emissions, renewables, and efficiency indicators across 170+ countries. To enhance LLMs' analytical capabilities for energy transition assessment, we incorporated the IEA\_Energy\_Dataset \cite{ZihaoLi2023IEAEnergyDataset} with 358,466 data points into our training corpus.

\begin{table}[htbp]
\centering
\caption{Text Corpus Statistics for Helios Training.}
\label{tab:corpus_stats}
\resizebox{1.0\linewidth}{!}{%
\begin{tabular}{l|c|c|c}
\toprule
\multirow{2}{*}{\textbf{Data Source}} & \textbf{Smart Energy Corpus} & \multicolumn{2}{c}{\textbf{EnerBase}} \\
\cmidrule(lr){2-2} \cmidrule(lr){3-4}
 & \textbf{Documents} & \textbf{Documents} & \textbf{Tokens (B)} \\
\midrule
OAP & 173,541 & 153,165 & 2.314 \\
OAJP & 32,459 & 30,249 & 0.57\\
SP & 363 & 342 & 0.038 \\
DMT\&AC & 5,667 & 4,039 & 0.015 \\
IEAD & 358,466 & 345,874 & 0.019 \\
\midrule
\textbf{Total} & \textbf{570,496} & \textbf{533,669} & \textbf{2.956} \\
\bottomrule
\end{tabular}%
}
\vspace{-0.5mm}
\begin{minipage}{\textwidth}

\end{minipage}
\end{table}

 \subsubsection{Smart Energy Corpus Processing}
\textbf{PDF Convert.} Collected corpus primarily exists in PDF format, necessitating conversion to a unified format suitable for model training. Scientific literature contains abundant structured information, including tables, equations, and formulas; direct conversion to TXT format would result in critical information loss, causing LLMs to learn incomplete or incorrect content. Therefore, we selected Markdown as our unified conversion format to preserve these essential structural elements.

\begin{figure*}[!h]
  \centering
   \includegraphics[width=1.0\linewidth]{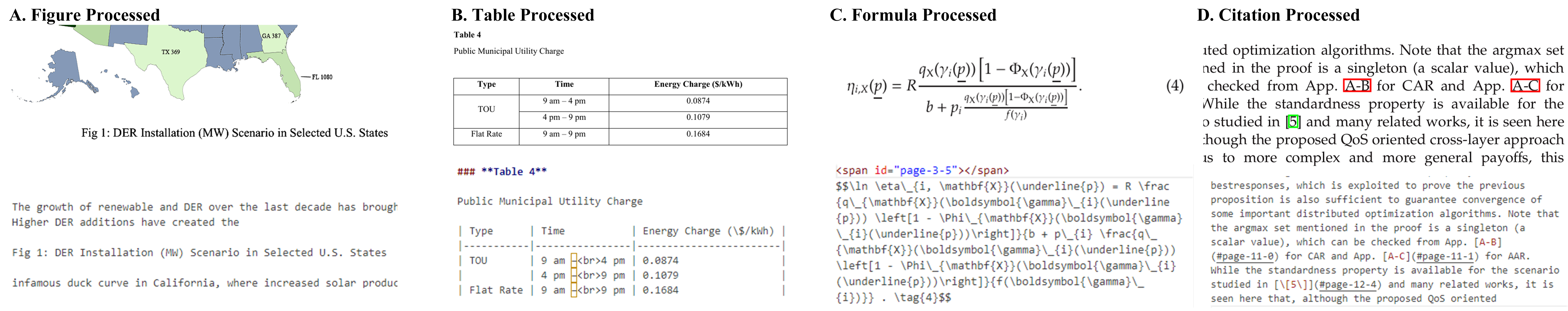}
   \caption{ Text processed by the Parsing-Agent. A.Images: only the captions are retained, image bodies are removed; B.Tables: converted to Markdown format; C.Complex mathematical formulae: converted to Markdown format; D.Citations: for each citation, the corresponding page numbers of the referenced literature are specified.} 
   \label{fig:md}
   \vspace{-2mm}
\end{figure*}

To balance computational throughput with structural integrity, we developed Python scripts based on Marker \cite{paruchuri2025marker}. For processing efficiency, we deployed 10 servers equipped with NVIDIA RTX 4090 GPUs in a distributed architecture, each server configured with six parallel conversion workers. To enhance quality, we integrated OpenAI's GPT-4o as an intelligent agent (Parsing-Agent) to perform table reconstruction, mathematical formula standardization, form parsing, figure description generation, and reference normalization, ensuring structural completeness. Detailed hyperparameter configurations are provided in the supplementary table. Our system achieved an average processing speed of 2.21 seconds per page, completing the entire conversion process within 5 days. Fig. \ref{fig:md} demonstrates sample conversion results. The computationally efficient and structurally complete PDF-to-Markdown conversion framework, based on intelligent agents, presented in this paper, has been open-sourced on GitHub along with the dataset.

\textbf{Content Filtering.} Building upon this foundation, we additionally implement filtering mechanisms to remove private information, harmful content, and unintelligible or corrupted text.

\textbf{Deduplication.} Nevertheless, the Smart Energy Corpus inevitably contains a proportion of semantically similar fragments, causing the model during pre-training to update along nearly identical gradient directions and thus to "memorise" specific passages rather than acquire generalisable logical patterns \cite{tirumala2023d4}. To address this problem, following the methodology outlined in \cite{abbas2023semdedup}, we developed an efficient large-scale deduplication agent, Corpus Distiller, built on BERT-base. Corpus Distiller first performs K-Means clustering in the embedding space and subsequently removes samples located within the same epsilon-ball in each cluster. 


\begin{table}[htbp]
\caption{Datasets used to train Helios during the Universal Human Instruction Comprehension phase.}
\label{tab:UHIC_d}
\centering

{\fontsize{9pt}{10pt}\selectfont
\setlength{\tabcolsep}{2pt} 
\begin{tabularx}{\columnwidth}{@{}X r@{}}
\toprule
\textbf{Dataset} & \textbf{Prompts} \\
\midrule
Alpaca-cleaned \cite{alpaca}       & 51\,800 \\
Dolly-15K \cite{DatabricksBlog2023DollyV2}            & 15\,011 \\
Natural-Instructions \cite{MuennighoffNaturalInstructions} & 30\,000 \\
python\_code\_25k \cite{FlytechPythonCodes25k}     & 24\,813 \\
OpenR1-Math-220k \cite{OpenR1Math220k}      & 28\,120 \\
Toolbench \cite{qin2023toolllm}            & 10\,328 \\
\midrule
\textbf{Total}        & \textbf{160\,072} \\
\bottomrule
\vspace{-8mm}
\end{tabularx}
}

\end{table}
\vspace{-4mm}

\subsection{Instruction Tuning Data}
Instruction Tuning is the key to bridging large‑scale unsupervised pre‑trained models with downstream applications. We have constructed a two‑phase instruction fine‑tuning framework of "Universal Human Instruction Comprehension (UHIC) to Domain-specific Task Adaptation (DS-TA)": First, high‑quality general instruction samples are employed to conduct preliminary fine‑tuning, enabling the model to learn to accomplish tasks according to natural‑language instructions; subsequently, knowledge‑intensive, specialized data are introduced for further fine‑tuning, thereby enhancing the model’s adaptability to domain‑specific tasks. For these two phases, we curate a complementary general instruction dataset and knowledge‑intensive dataset \textbf{EnerInstruct}, each uniformly organized in an <instruction,input,output> triplet format.
\subsubsection{Universal Human Instruction Comprehension Data}
In this stage, we have carefully selected six highly-recognized and high-quality open-source general-purpose supervised datasets: Alpaca-cleaned \cite{alpaca}, Dolly-15K \cite{DatabricksBlog2023DollyV2}, Natural-Instructions \cite{MuennighoffNaturalInstructions,naturalinstructions,supernaturalinstructions}, python\_code\_25k \cite{FlytechPythonCodes25k}, OpenR1-Math-220k \cite{OpenR1Math220k}, and Toolbench \cite{qin2023toolllm}. These datasets cover universal instruction understanding, mathematical reasoning, code enhancement, and tool utilization domains to improve Helios's foundational capabilities and domain application potential. The data volume of each dataset in the UHIC phase is shown in Table \ref{tab:UHIC_d}.
\subsubsection{Domain-specific Task Adaptation data: EnerInstruct}
\textbf{Seed Data Collection.} In this study, we engaged 10 senior experts in the smart energy domain to manually construct sample pairs for eleven downstream tasks: Fact Verification (FV), Reasoning (Res),  Named Entity Recognition (NER), Summarization (Sum), Word Semantics (WS), Question and Answers (Q\&A), Text Classification (TC), Explanation (Exp), Energy System Modeling (ESM), Single‑Choice (S-C) and Multiple-Choice (M-C). Which across fourteen sub-fields: clean energy, cogeneration, combined cooling–heating–and–power, distributed energy, energy hub, energy management system, energy optimization, energy storage, energy transition, integrated energy, load forecasting, smart energy, smart grid, and virtual power plant. The resulting seed dataset, covers 14 sub-fields and 10 task categories, comprising 10\,000 samples.

\begin{table}[t]  
  \caption{Statistics of EnerInstruct categorized by tasks.}
  \label{tab:task_stats}  
  \resizebox{1.0\linewidth}{!}{%
  \begin{tabular}{c|c|cc|c}
    \toprule
    \multirow{2}{*}{\textbf{Tasks}} & \multirow{2}{*}{\textbf{Records}} & \multicolumn{2}{|c|}{\textbf{Dataset Quality Optimization}} &  \multirow{2}{*}{\textbf{Total (Cleaned)}} \\
    \cmidrule(lr){3-4}
    & & Filtered & optimized &   \\
    \midrule
    FV & 20,839 & 17,370 & 706 & 4,175   \\
    Res & 6,057  & 351 & 100 & 5,806   \\
    NER &  423 & 327 & 283 & 379  \\
    Sum & 449 & 392 & 323 & 380    \\
    WS    & 6,830 & 6,166 & 5,714 & 6,378   \\
    Q\&A  & 11,973  & 7,900 & 3,878 & 7,951    \\
    TC    & 5,486   & 1,513  & 648 & 4,621  \\
    Exp     & 9,003 & 1,785  & 1,045 & 8,263    \\
    ESM     & 721 & 674  & 672 & 719  \\
    S-C      & 8,234 & 2,638  & 1,523 & 7,119   \\
    M-C      & 10,780 & 3,368  & 2,213 & 9,625   \\
    \midrule
    \textbf{Entire Data} & \textbf{80,795} & \textbf{42,484} & \textbf{17,105} & \textbf{55,416}\\
    \bottomrule
  \end{tabular}%
  }
\end{table}

\textbf{Dataset Augmentation.} Smart energy encompasses multiple subfields, each exhibiting unique statistical characteristics and potential patterns. To ensure the professionalism and accuracy of the generated results, we design domain-specific expert agents for each subfield, enabling them to independently generate high-quality sample pairs for their respective areas and achieve parallelization and high-throughput data output. Specifically, we first refine the literature from each subfield within the Open-access Authoritative Journal Papers using a two-stage selection method based on "local citation count" and "co-citation analysis" to identify high academic value papers that constitute the foundational knowledge and theoretical framework of the discipline. These papers serve as a high-quality corpus (using the energy storage subfield as an example, refer to Algorithm \ref{alg:refinement}). Subsequently, we fine-tune the corresponding expert agents using seed datasets from each subfield, enabling them to autonomously generate <instruction, input, output> triplets that conform to training standards based on the high-quality corpus. The original DS-TA phase data are constructed, with task assignment details provided in Table \ref{tab:task_stats}.
\begin{algorithm}[t]
\caption{\small Two-Stage Literature Refinement for Energy Storage Domain}
\label{alg:refinement}
\scriptsize   
\begin{algorithmic}[1]
\Require $P$: Publication set; $\theta_{LC}$: Citation threshold (70-th percentile); $\varepsilon$: DBSCAN distance (0.7); $\text{MinPts}$: DBSCAN density (5); $m_k$: Top papers per cluster
\Ensure $V''$: Refined core paper set

\State \textbf{Stage 1: Local Citation Filtering}
\State Construct paper network $V=\{v_1,...,v_n\}$ from $P$ where each $v_i$ represents a paper
\State Define citation indicator: $I(v_i\rightarrow v_j)=1$ if paper $v_i$ cites paper $v_j$, 0 otherwise
\For{$v_i \in V$}
    \State $LC(v_i) \leftarrow \sum_{v_j \in V} I(v_j \rightarrow v_i)$ \Comment{Local citation count}
\EndFor
\State $V' \leftarrow \{v_i \mid LC(v_i) \geq \theta_{LC}\}$ \Comment{Filter high-cited papers}

\State \textbf{Stage 2: Co-citation Analysis}
\State Build co-citation matrix $c_{ij} = \sum_{v_k \in V} I(v_k \rightarrow v_i) I(v_k \rightarrow v_j)$
\State $s_{ij} \leftarrow c_{ij}/\sqrt{c_{ii} c_{jj}}$ \Comment{Normalized co-citation similarity}
\State $\{C_1, \ldots, C_K\} \leftarrow \textsc{DBSCAN}(S, \varepsilon, \text{MinPts})$ \Comment{Cluster by similarity matrix $S$}
\For{each cluster $C_k$}
    \For{$v_i \in C_k$}
        \State $CD(v_i) \leftarrow \sum_{v_j \in C_k} s_{ij}$ \Comment{Centrality degree within cluster}
    \EndFor
    \State $T_k \leftarrow$ top-$m_k$ papers in $C_k$ ranked by $CD(v_i)$
\EndFor
\State $V'' \leftarrow \bigcup_{k=1}^{K} T_k$ \Comment{Union of top papers from all clusters}
\end{algorithmic}

\vspace{1mm}
\tiny   
\textit{Note: For the energy storage domain, $|P|=5204$, $|V'|=1561$, and $|V''|=312$. 
We target approximately 300 papers per domain by adjusting $m_k$ (5–15). 
Statistics for other domains are given in Appendix~D.}
\end{algorithm}

\textbf{Dataset Quality Optimization.} During dataset construction, domain‑expert agents generated a large number of highly specialised samples for the various tasks. Nevertheless, the stochastic nature of sampling, the structural constraints imposed by the context‑window length, and the potential hallucinations produced by LLMs can all cause fluctuations in sample quality and completeness. Extensive empirical work demonstrates that dataset size governs coverage and diversity, whereas sample quality determines the attainable upper bound on performance; the two must be carefully balanced. Suppose low‑quality samples containing redundancy, noise, or inconsistent annotations are used for training. They will dilute the informative signal, amplify systemic bias, and ultimately erode the model’s ability to follow instructions. To this end, we constructed a Check-Agent based on OpenAI o1, categorized by task, scoring each sample across the four dimensions of accuracy, completeness, relevance, and usefulness (out of 10), and providing reasons.

\begin{figure*}[!h]
  \centering
   \includegraphics[width=1.0\linewidth]{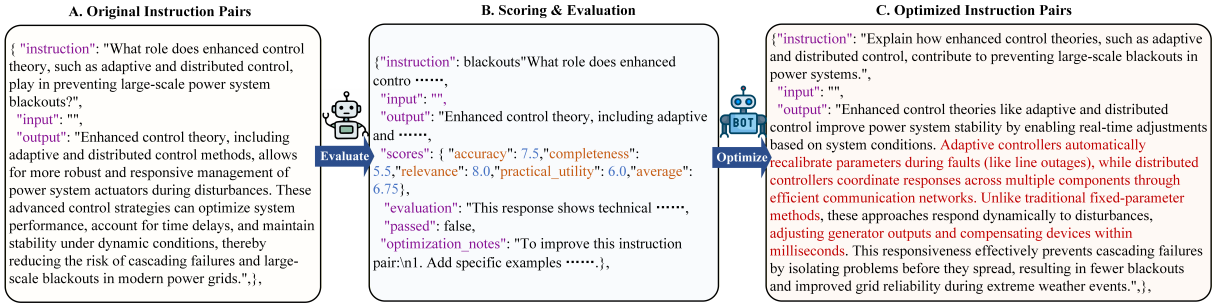}
   \vspace{-2mm}
   \caption{Dataset Quality Optimization workflow example. (A) Text before processing; (B) the Check‑Agent scores the text quality and provides optimization suggestions; (C) the Optimization‑Agent generates the optimized text based on those suggestions. We mark the differences in Red.} 
   \label{fig:Evaluate}
\end{figure*}

Samples that reach or exceed the threshold are retained, whereas those that do not are forwarded to an independent Optimization-Agent. Guided by the evaluation feedback, this agent performs automatic remediation—correcting errors, supplementing and enriching content, or conducting deeper analysis as necessary. The revised sample is then returned to the Check-Agent for re‑evaluation. This “scoring–Optimization–re‑scoring” loop may iterate up to ten times: if a sample passes within the allotted rounds, it is admitted to the training corpus; if it fails all ten rounds, it is deemed irreparable and permanently discarded. Check‑Agent and the Optimization-Agent collaboratively optimise the data workflow, as shown in Fig \ref{fig:Evaluate}. This procedure ultimately yields dataset \textbf{H‑EnerInstruct}.

\textbf{Expert Manual Validation.} Finally, a panel of 12 domain experts rigorously examined each task sample in \textbf{H‑EnerInstruct} (sampling 100–200 entries per task, proportional to that task’s size). Tasks that did not meet the required standard were flagged, and the experts issued uniform revision guidelines that were then refined by OpenAI o1 to ensure the dataset’s reliability. The optimized data were merged with the seed dataset to produce the final DS-TA phase dataset, \textbf{EnerInstruct} (Table \ref{tab:task_stats}). The statistics on expert optimization iterations are reported in Supplementary Section C.

\subsection{RLHF Data: EnerReinforce}
After large‑scale pre‑training and supervised instruction fine‑tuning, Helios can already address a wide range of tasks in the energy domain. Nevertheless, these stages seldom make human values or preferences explicit, so the resulting models may acquire generation patterns that diverge from human expectations. To align Helios more effectively with human preferences, we adopt a targeted, two‑stage approach consisting of reward model training followed by rejection sampling fine‑tuning, and constructed \textbf{EnerReinforce}, which includes:

\textbf{1) Reward Model Training Data:}
We sampled 5,000 subjective questions $Q_{RM} = \{ q_i \}_{i=1}^{5000}$,  and their expert answers $E_{Exp} = \{ e_i \}_{i=1}^{5000}$ from seed dataset. For each $q_i$, additional answers $E_{WLH}$, $E_{GPT-4.0}$, $E_{GPT-3.5}$ were generated with (i) a Write‑like‑Human agent, (ii) GPT‑4.0, and (iii) GPT‑3.5. The four answers were then ranked by quality in the order $E_{Exp}$ > $E_{WLH}$ > $E_{GPT-4.0}$ > $E_{GPT-3.5}$, and pair adjacent response to obtain 3 sets of positive and negative sample pairs $\mathcal{P}_{i} = \bigl\{ \bigl(E_{Exp},\, E_{WLH}\bigr),\; \bigl(E_{WLH},\, E_{GPT-4.0}\bigr),\;\allowbreak \bigl(E_{GPT-4.0},\, E_{GPT-3.5}\bigr) \bigr\}$, and formed the Reward‑model training dataset $\mathcal{X}_{RM} = \{\mathcal{P}_{1}, \mathcal{P}_{2}, …,\mathcal{P}_{5000}\}$.

\textbf{2) Rejection Sampling Fine‑tuning Data:}
From the subjective part of \textbf{EnerInstruct}, we selected 10,000 questions $Q_{RS} = \{ q_i \}_{i=1}^{10000}$, that do not overlap with the seed set. Helios generated five candidate answers $A_i = \{ a_i^1, a_i^2, …, a_i^5 \}$ for each question $q_i$, and these candidates  serve as the basis for the  rejection‑sampling fine‑tuning stage $\mathcal{X}_{RS} = \{ (q_1, A_1),(q_2, A_2), …,(q_{10000}, A_{10000})\}$.

\subsection{Evaluation on Expertise in Smart Energy: EnerBench}
To systematically assess the problem-solving capabilities of LLMs on scientific questions in smart energy research, we developed EnerBench, whose item-generation workflow adheres to a dual-track paradigm of Public-bank retrieval and Expert-directed authoring:

1) \textbf{Public-bank Retrieval:} Using each sub-discipline as a search keyword, representative questions were automatically harvested from multiple open-source evaluation platforms, ensuring extensive topical coverage and diversity.

2) \textbf{Expert-directed Authoring:} Five senior scholars in the smart energy domain were commissioned to craft additional, high-quality items for every task within each sub-discipline, thereby augmenting the benchmark’s novelty and difficulty.

In its final form, EnerBench comprises 887 objective questions (S-C, M-C, and FV) and 625 subjective questions (Q\&A, Exp, and ESM). The detailed distribution of questions across sub-disciplines is provided in Table \ref{tab:bench}.

\begin{table}[htbp]
\caption{The statistics of EnerBench.}
\label{tab:bench}
\centering

{\fontsize{9pt}{10pt}\selectfont
\setlength{\tabcolsep}{4pt} 

\begin{tabularx}{\columnwidth}{@{}p{2.6cm} X r@{}}
\toprule
\textbf{Question Type} & \textbf{Task} & \textbf{Prompts} \\
\midrule
\multirow{3}{*}{Objective task} & S-C       & 405 \\
                                & M-C     & 254 \\
                                & FV   & 228 \\
\midrule
\multirow{3}{*}{Subjective tasks} & Q\&A       & 196 \\
                                  & Exp           & 249 \\
                                  & ESM     & 180 \\
\bottomrule
\end{tabularx}
}

\end{table}
\vspace{-6pt}



\begin{table*}[htbp]
\centering
\vspace{-2mm}
\caption{Comparison of the performance of different LLMs across all tasks in EnerBench. The best results are indicated in \textbf{bold}, and the second-best results are \underline{underlined}.}
\label{tab:result}
\resizebox{0.8\linewidth}{!}{%
\begin{tabular}{l|c|c|c|ccc|ccc|ccc}
\toprule
\multirow{2}{*}{\textbf{Model}} &
\multirow{2}{*}{\textbf{S-C}} &
\multirow{2}{*}{\textbf{M-C}} &
\multirow{2}{*}{\textbf{FV}} &
\multicolumn{3}{c|}{\textbf{ESM}} &
\multicolumn{3}{c|}{\textbf{Exp}} &
\multicolumn{3}{c}{\textbf{Q\&A}} \\ 
\cmidrule(lr){5-7}\cmidrule(lr){8-10}\cmidrule(lr){11-13}
 &  &  &  & \textbf{A} & \textbf{E} & \textbf{H} & \textbf{A} & \textbf{E} & \textbf{H} & \textbf{A} & \textbf{E} & \textbf{H} \\
\midrule
Qwen3-8B-Instruct          & 50.24\% & 27.56\% & 47.81\% & 1.74 & 5.88 & D & 1.63 & 5.04 & D & 4.74 & 6.50 & C \\
Llama3-8B-Instruct           &  68.42\% &  37.60\% &   58.77\% & 3.29   & 6.03  & D  & 3.53  & 6.29  & D  & 5.13  & 6.47  & C  \\
Qwen3-14B-Instruct        & 64.59\% & 35.24\% & 54.61\% & 2.26  & 6.54  & D  & 4.36  & 6.90  & C  & 6.22  & 7.06  & C  \\
Qwen3-32B-Instruct         & 80.14\% & 44.09\% & 62.72\% & 3.82  &  6.93 & D  & 5.51  &  7.32 & C  & 6.83  & 7.51  & \textbf{B}  \\
GPT-3.5-Turbo                & 91.63\% &  \underline{53.93\%} & 84.65\% & \underline{6.03}  & \underline{8.05}  &  \underline{C}  & 6.94  & 8.57  & \textbf{B}  & 7.24  & \underline{8.37}  & \textbf{B}  \\
GPT-4                      & \textbf{95.69\%} & \textbf{61.18\%} & \textbf{93.86\%} & \textbf{7.61}  & \textbf{8.97}  & \textbf{B}  & \textbf{8.63}  & \textbf{9.58}  & \textbf{B}  & \textbf{7.64}  & \textbf{9.21}  & \textbf{B}  \\
\textbf{Helios}               & \underline{93.78\%} & 53.58\% & \underline{89.91\%} & 5.73 & 7.83 & \underline{C} & \underline{7.03} & \underline{9.19} & \textbf{B} & \underline{7.39} & 8.26 & \textbf{B} \\
\bottomrule
\end{tabular}%
}
\end{table*}

\section{Helios training settings}


\subsection{Pre-training}
During the pre-training stage, we employ the Qwen-2.5 7B model \cite{yang2024qwen2} (7.62B trainable parameters) as the initialization weights for Helios. A single-epoch training is subsequently conducted on a domain-specific corpus of approximately 3 billion tokens in the smart energy domain (22532 gradient update steps); the training hardware configuration consists of four NVIDIA A100-SXM 80 GB GPUs, with a total training time of 87 hours. The principal hyper-parameter settings are as follows: a peak learning rate of 3e-5, global batch size of 64, and a corresponding micro-batch size of 2.
\subsection{Instruction tuning}
In both stages of instruction learning (UHIC and DS-TA), we employ the Low-Rank Adaptation (LoRA) technique: while keeping the pre-trained weights \(W_0 \in \mathbb{R}^{n \times d}\) completely frozen, we inject two trainable low-rank matrices \(A \in \mathbb{R}^{n \times r}\) and \(B \in \mathbb{R}^{r \times d}\) in parallel (\(r \ll \min(n,d)\)). Here, \(n\) and \(d\) represent the input and output dimensions of the weight matrix \(W_0\), and \(r\) denotes the rank of the low-rank matrices. This approach preserves the general representations learned from large-scale corpora during pre-training, while significantly reducing the number of trainable parameters and lowering computational costs. The corresponding forward propagation is given by:
\begin{equation}
h = W_0 x + BAx,
\end{equation}

where \( h \) denotes the adapted output. The training hardware configuration consists of four NVIDIA RTX 4090 GPUs, with a total training time of 17 hours. During the instruction tuning stage, a two-stage fine-tuning of the model was performed. The model was first fine-tuned with generic instructions and then fine-tuned with knowledge enhancement. In the UHIC stage, the key hyperparameter settings are as follows: a peak learning rate of 2e-5, global batch size of 64, and a corresponding micro-batch size of 2. In the DS-TA stage, the key hyperparameter settings are as follows: a peak learning rate of 1e-5, a global batch size of 64, and a corresponding micro-batch size of 2.

\subsection{RLHF}
\textbf{Reward Model Training.}
We employ a pairwise ranking loss to train reward model, enabling it to distinguish between responses of varying quality:
\vspace{-4mm}

\begin{small}
\begin{equation}
\label{eq:rm-loss}
\mathcal{L}_{\mathrm{RM}}
 = -\frac{1}{\lvert \mathcal{D}_{\mathrm{RM}} \rvert}\!
   \sum_{i=1}^{\mathcal{D}_{\mathrm{q}}}
   \sum_{j=1}^{\mathcal{D}_{\mathrm{pair}}}
   \log \sigma\!\bigl( r_{\phi}(q_i,a_i^{j+}) - r_{\phi}(q_i,a_i^{j-}) \bigr),
\end{equation}
\end{small}

where $ \mathcal{D}_{RM} $ denotes the set of training examples ($ \mathcal{D}_{RM} = \mathcal{D}_q *\mathcal{D}_{\mathrm{pair}}$), $ \mathcal{D}_q$ denotes the cardinality of $Q_{RM}$, $\mathcal{D}_{\mathrm{pair}}$ denotes the number of positive--negative sample pairs associated with $q_i$. $a_i^{j+}$ and $a_i^{j-}$ represent the $j$-th positive and negative samples of $q_i$, respectively. $ r_{\phi}(q_i,a_i^{j}) $ is the quality score assigned by the reward model to response $ a_i^{j} $; and $\sigma(\cdot) $ is the sigmoid function, which maps the difference in scores to the probability that the positive sample is preferred over the negative one. By minimizing $ \mathcal{L}_{\mathrm{RM}} $, the model is driven to enlarge the gap between $ r_{\phi}(x, y^{+}) $ and $ r_{\phi}(x, y^{-}) $, thereby learning to distinguish responses of differing quality. For the hyperparameters, we train for three epochs with a batch size of 8, and the warm‑up stage accounts for 5\% of the total steps.

\textbf{Rejection Sampling Fine-tuning.}
During the Rejection Sampling fine-tuning phase, the reward model is used to evaluate and rank $\mathcal{X}_{RS}$:
\begin{equation}
s_i = \{r_{\phi}(q_i, a_i^j)\}_{j=1}^{\mathcal{D}_{c}},A_i^{\ast} = sort(A_i, desc\ by\  s_i),
\end{equation}

$\mathcal{D}_c$ denotes the number of candidate responses in  $A_i$, $s_i$ denotes the score assigned to each response by the reward model. $ A_i^*$ is obtained by sorting $A_i$  in descending order of $s_i$. Then, select the Top‑k samples as the “gold standard” for further fine‑tuning Helios:
\begin{equation}
\mathcal{X}_{\text{RS}}^{Gold} = \left\{ (q_i, a_i^{\ast}) |q_i \in Q_{RS}, a_i^{\ast} \in \text{TopK}(A_i^{\ast}, k) \right\}_{i=1}^{\mathcal{D}_r},
\end{equation}

where $\mathcal{D}_r$ denotes the number of questions in $Q_{RS}$, $a_i^{\ast}$ is the set of the top  k values sampled from $A_i^{\ast}$. 
We trained the model for 5 epochs with a learning rate of 3e-5 and a batch size of 64.

\section{Evaluation and Results}
We evaluated the performance of Helios, Qwen3-8B-Instruct, Llama3-8B-Instruct, Qwen3-14B-Instruct, Qwen3-32B-Instruct, GPT3.5-Turbo and GPT-4 on EnerBench and compared their results. The results are presented in Table \ref{tab:result}.


\textbf{Objective Tasks in EnerBench.}
For objective tasks, performance is evaluated using accuracy. Specifically, for multiple-choice items, the scoring rubric is: full credit is awarded only when all correct options are selected; partial credit is granted when some correct options are omitted; and no credit is given if any incorrect option is chosen. Helios attains an average accuracy of 79.09\% in answering object questions, markedly outperforming models of comparable size such as Qwen3-8B-Instruct (41.87\%) and Llama3-8B-Instruct (54.93\%), and reaching a level comparable to GPT-4 with approximately 220 billion parameters. This indicates that the model successfully acquired intelligent-energy domain knowledge during further pre-training.

\textbf{Subjective Tasks in EnerBench.}
For subjective tasks, we implemented a tri-dimensional evaluation framework: A-Score (GPT-o1 benchmark-based comparative assessment on a 10-point scale), E-Score (GPT-o1 independent quality assessment on a 10-point scale), and H-Grade (expert evaluation using an A/B/C/D grading system). The assessment results demonstrate that Helios outperforms parameter-equivalent models like Qwen3-8B-Instruct and Llama3-8B-Instruct across domain-specific QA, Exp, and ESM tasks. Specifically, Helios approaches GPT-4 capability levels in QA and Exp tasks. Regarding ESM capabilities, Helios can leverage energy domain-specific libraries for complex problem modeling. However, it still exhibits a performance gap compared to GPT-4 due to parameter size constraints, yet achieves performance comparable to GPT-3.5-Turbo. We provide a detailed discussion of model hallucinations in section H of the supplementary materials.

\begin{figure}[t]
    \centering
    \includegraphics[width=0.95\linewidth]{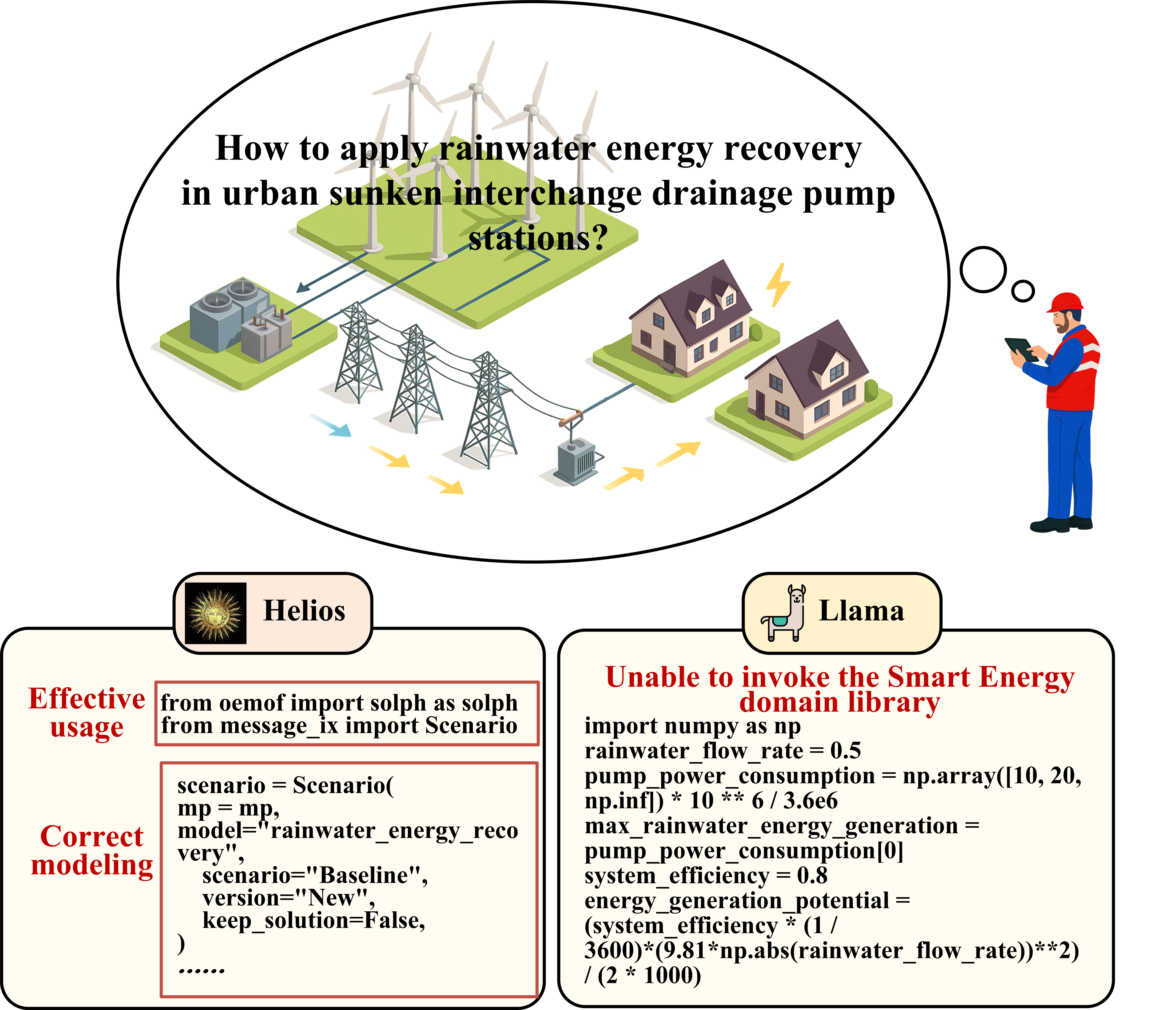}
    \caption{ Case analysis of modeling tasks in the smart energy domain.}
    \label{fig:code}
    \vspace{-4mm}
\end{figure}
\textbf{Exploring the Potential of Helios.} We attempt to address a practical modelling and Optimization task in the smart energy domain using Helios. In this example, our requirement is: \textit{How to apply rainwater energy recovery in urban sunken interchange drainage pump stations?} and to provide the implementation code (Fig. \ref{fig:code}). It can be observed that Helios is able to effectively invoke domain-specific packages for intelligent energy (oemof and message\_ix) to accomplish the modelling task, whereas Llama3-8B can only call numpy to perform purely numerical computations, which deviates substantially from the task requirements and lacks practical relevance to the energy domain.

\section{Conclusion}

In this study, we introduce Helios, the first LLM explicitly developed for the smart energy domain, capable of addressing diverse tasks. We introduce EnerSys, a comprehensive multi-agent pipeline that furnishes Helios with high-quality data, producing EnerBase, EnerInstruct, and EnerReinforce. We also release Benchmark, the domain’s first evaluation suite, enabling systematic appraisal of LLMs on smart energy tasks. Experiments show that, Helios offers significant gains in domain knowledge and task performance. 



\section*{Limitations}

Although Helios has demonstrated excellent capabilities in knowledge integration and automatic code generation within the smart energy domain, its role is consistently positioned as an "intelligent reference assistant" rather than an autonomous decision-making engine. In high-risk tasks such as power system modeling, dispatch, and safety assessment, Helios only outputs code drafts and inferential suggestions for review by engineers. Direct deployment without professional review could lead to significant economic losses or even physical risks due to potential model assumption biases, numerical instability, or omission of boundary conditions. Consequently, the model's outputs do not constitute an engineering guarantee. The final decision-making responsibility must be borne by the user and their affiliated institution; when results are uncertain or contradict engineering experience, it is essential to revert to traditional manual calculation and simulation for verification.

 Concerning hallucinations in Helios, they mainly manifest as linguistic repetition, instruction misunderstanding, conceptual confusion, and structural errors. For instance, in factual judgment tasks, the model occasionally misinterprets the task as question-answering, a problem that is significantly mitigated by explicitly appending "Please output True/False directly" to the prompt. Conceptual confusion stems from the interdisciplinary, fragmented, and rapidly evolving nature of smart energy knowledge; experiments show its occurrence rate remains within an acceptable range. Linguistic repetition and structural hallucinations are largely associated with the base model, Qwen-2.5-7B, and are difficult to eliminate completely through domain-specific fine-tuning alone; thus, they are not a primary focus of this paper.In summary, Helios has the aforementioned limitations regarding ethics, risks, and deployment, and should be applied cautiously within a strict framework of human-computer collaboration and safety governance.

\section*{Ethical considerations}
Regarding ethics and bias, Helios is primarily trained on high-quality corpora such as academic papers and monographs, and its instruction data has undergone rigorous cleaning, resulting in minimal potential for ethical or bias issues.

\section*{Acknowledgments}
This work is supported by the National Natural Science Foundation of China (Grant No.52576234) and the National Key R \& D Program of China (Grant No.2023YFE0108600). This work is also supported by the "Pioneer" and "Leading Goose" R \& D Program of Zhejiang (Grant No. 2025C02237)



\bibliography{custom}

\begin{thebibliography}{59}
\providecommand{\natexlab}[1]{#1}

\bibitem[{Abbas et~al.(2023)Abbas, Tirumala, Simig, Ganguli, and Morcos}]{abbas2023semdedup}
Amro Abbas, Kushal Tirumala, D{\'a}niel Simig, Surya Ganguli, and Ari~S Morcos. 2023.
\newblock \href {https://arxiv.org/abs/2303.09540} {Semdedup: Data-efficient learning at web-scale through semantic deduplication}.
\newblock \emph{arXiv preprint arXiv:2303.09540}.

\bibitem[{Amirkhani and Barshooi(2022)}]{amirkhani2022consensus}
Abdollah Amirkhani and Amir~Hossein Barshooi. 2022.
\newblock \href {https://link.springer.com/article/10.1007/s10462-021-10097-x} {Consensus in multi-agent systems: a review}.
\newblock \emph{Artificial Intelligence Review}, 55(5):3897--3935.

\bibitem[{Bai et~al.(2023)Bai, Bai, Chu, Cui, Dang, Deng, Fan, Ge, Han, Huang, Hui, Ji, Li, Lin, Lin, Liu, Liu, Lu, Lu, Ma, Men, Ren, Ren, Tan, Tan, Tu, Wang, Wang, Wang, Wu, Xu, Xu, Yang, Yang, Yang, Yang, Yao, Yu, Yuan, Yuan, Zhang, Zhang, Zhang, Zhang, Zhou, Zhou, Zhou, and Zhu}]{bai2023qwen}
Jinze Bai, Shuai Bai, Yunfei Chu, Zeyu Cui, Kai Dang, Xiaodong Deng, Yang Fan, Wenbin Ge, Yu~Han, Fei Huang, Binyuan Hui, Luo Ji, Mei Li, Junyang Lin, Runji Lin, Dayiheng Liu, Gao Liu, Chengqiang Lu, Keming Lu, and 29 others. 2023.
\newblock \href {https://arxiv.org/abs/2309.16609} {Qwen technical report}.
\newblock \emph{Preprint}, arXiv:2309.16609.

\bibitem[{Bai et~al.(2025)Bai, Chen, Liu, Wang, Ge, Song, Dang, Wang, Wang, Tang, Zhong, Zhu, Yang, Li, Wan, Wang, Ding, Fu, Xu, Ye, Zhang, Xie, Cheng, Zhang, Yang, Xu, and Lin}]{bai2025qwen2}
Shuai Bai, Keqin Chen, Xuejing Liu, Jialin Wang, Wenbin Ge, Sibo Song, Kai Dang, Peng Wang, Shijie Wang, Jun Tang, Humen Zhong, Yuanzhi Zhu, Mingkun Yang, Zhaohai Li, Jianqiang Wan, Pengfei Wang, Wei Ding, Zheren Fu, Yiheng Xu, and 8 others. 2025.
\newblock \href {https://arxiv.org/abs/2502.13923} {Qwen2.5-vl technical report}.
\newblock \emph{Preprint}, arXiv:2502.13923.

\bibitem[{Bi et~al.(2024)Bi, Zhang, Xue, Ou, Ji, Zheng, and Chen}]{bi-etal-2024-oceangpt}
Zhen Bi, Ningyu Zhang, Yida Xue, Yixin Ou, Daxiong Ji, Guozhou Zheng, and Huajun Chen. 2024.
\newblock \href {https://doi.org/10.18653/v1/2024.acl-long.184} {{O}cean{GPT}: A large language model for ocean science tasks}.
\newblock In \emph{Proceedings of the 62nd Annual Meeting of the Association for Computational Linguistics (Volume 1: Long Papers)}, pages 3357--3372, Bangkok, Thailand. Association for Computational Linguistics.

\bibitem[{Ceglia et~al.(2020)Ceglia, Esposito, Marrasso, and Sasso}]{ceglia2020smart}
F~Ceglia, P~Esposito, E~Marrasso, and M~Sasso. 2020.
\newblock \href {https://www.sciencedirect.com/science/article/pii/S0959652620301657} {From smart energy community to smart energy municipalities: Literature review, agendas and pathways}.
\newblock \emph{Journal of Cleaner Production}, 254:120118.

\bibitem[{Chen et~al.(2024)Chen, Yan, Liu, Yan, and Tjernberg}]{chen2024novel}
Fuhao Chen, Jie Yan, Yongqian Liu, Yamin Yan, and Lina~Bertling Tjernberg. 2024.
\newblock \href {https://www.sciencedirect.com/science/article/pii/S0306261924002216} {A novel meta-learning approach for few-shot short-term wind power forecasting}.
\newblock \emph{Applied Energy}, 362:122838.

\bibitem[{Chen et~al.(2023)Chen, Su, Zuo, Yang, Yuan, Chan, Yu, Lu, Hung, Qian, Qin, Cong, Xie, Liu, Sun, and Zhou}]{chen2023agentverse}
Weize Chen, Yusheng Su, Jingwei Zuo, Cheng Yang, Chenfei Yuan, Chi-Min Chan, Heyang Yu, Yaxi Lu, Yi-Hsin Hung, Chen Qian, Yujia Qin, Xin Cong, Ruobing Xie, Zhiyuan Liu, Maosong Sun, and Jie Zhou. 2023.
\newblock \href {https://openreview.net/pdf?id=EHg5GDnyq1} {Agentverse: Facilitating multi-agent collaboration and exploring emergent behaviors}.
\newblock In \emph{The Twelfth International Conference on Learning Representations}.

\bibitem[{Chiang et~al.(2023)Chiang, Li, Lin, Sheng, Wu, Zhang, Zheng, Zhuang, Zhuang, Gonzalez, Stoica, and Xing}]{chiang2023vicuna}
Wei-Lin Chiang, Zhuohan Li, Zi~Lin, Ying Sheng, Zhanghao Wu, Hao Zhang, Lianmin Zheng, Siyuan Zhuang, Yonghao Zhuang, Joseph~E. Gonzalez, Ion Stoica, and Eric~P. Xing. 2023.
\newblock \href {https://lmsys.org/blog/2023-03-30-vicuna/} {Vicuna: An open-source chatbot impressing gpt-4 with 90\%* chatgpt quality}.

\bibitem[{Conover et~al.(2023{\natexlab{a}})Conover, Hayes, Mathur, Meng, Xie, Wan, Ghodsi, Wendell, and Zaharia}]{conover2023hello}
Mike Conover, Matt Hayes, Ankit Mathur, Xiangrui Meng, Jianwei Xie, Jun Wan, Ali Ghodsi, Patrick Wendell, and Matei Zaharia. 2023{\natexlab{a}}.
\newblock \href {https://www.databricks.com/blog/2023/03/24/hello-dolly-democratizing-magic-chatgpt-open-models.html} {Hello dolly: Democratizing the magic of chatgpt with open models}.

\bibitem[{Conover et~al.(2023{\natexlab{b}})Conover, Hayes, Mathur, Xie, Wan, Shah, Ghodsi, Wendell, Zaharia, and Xin}]{DatabricksBlog2023DollyV2}
Mike Conover, Matt Hayes, Ankit Mathur, Jianwei Xie, Jun Wan, Sam Shah, Ali Ghodsi, Patrick Wendell, Matei Zaharia, and Reynold Xin. 2023{\natexlab{b}}.
\newblock \href {https://www.databricks.com/blog/2023/04/12/dolly-first-open-commercially-viable-instruction-tuned-llm} {Free dolly: Introducing the world's first truly open instruction-tuned llm}.

\bibitem[{DeepSeek-AI et~al.(2024)DeepSeek-AI, Liu, Feng, Xue, Wang, Wu, Lu, Zhao, Deng, Zhang, Ruan, Dai, Guo, Yang, Chen, Ji, Li, Lin, Dai, Luo, Hao, Chen, Li, Zhang, Bao, Xu, Wang, Zhang, Ding, Xin, Gao, Li, Qu, Cai, Liang, Guo, Ni, Li, Wang, Chen, Chen, Yuan, Qiu, Li, Song, Dong, Hu, Gao, Guan, Huang, Yu, Wang, Zhang, Xu, Xia, Zhao, Wang, Zhang, Li, Wang, Zhang, Zhang, Tang, Li, Tian, Huang, Wang, Zhang, Wang, Zhu, Chen, Du, Chen, Jin, Ge, Zhang, Pan, Wang, Xu, Zhang, Chen, Li, Lu, Zhou, Chen, Wu, Ye, Ye, Ma, Wang, Zhou, Yu, Zhou, Pan, Wang, Yun, Pei, Sun, Xiao, Zeng, Zhao, An, Liu, Liang, Gao, Yu, Zhang, Li, Jin, Wang, Bi, Liu, Wang, Shen, Chen, Zhang, Chen, Nie, Sun, Wang, Cheng, Liu, Xie, Liu, Yu, Song, Shan, Zhou, Yang, Li, Su, Lin, Li, Wang, Wei, Zhu, Zhang, Xu, Xu, Huang, Li, Zhao, Sun, Li, Wang, Yu, Zheng, Zhang, Shi, Xiong, He, Tang, Piao, Wang, Tan, Ma, Liu, Guo, Wu, Ou, Zhu, Wang, Gong, Zou, He, Zha, Xiong, Ma, Yan, Luo, You, Liu, Zhou, Wu, Ren, Ren, Sha, Fu, Xu, Huang, Zhang, Xie, Zhang, Hao,
  Gou, Ma, Yan, Shao, Xu, Wu, Zhang, Li, Gu, Zhu, Liu, Li, Xie, Song, Gao, and Pan}]{liu2024deepseek}
DeepSeek-AI, Aixin Liu, Bei Feng, Bing Xue, Bingxuan Wang, Bochao Wu, Chengda Lu, Chenggang Zhao, Chengqi Deng, Chenyu Zhang, Chong Ruan, Damai Dai, Daya Guo, Dejian Yang, Deli Chen, Dongjie Ji, Erhang Li, Fangyun Lin, Fucong Dai, and 181 others. 2024.
\newblock \href {https://arxiv.org/abs/2412.19437} {Deepseek-v3 technical report}.
\newblock \emph{arXiv preprint arXiv:2412.19437}.

\bibitem[{Deng et~al.(2024)Deng, Zhang, He, Chen, Shi, Xu, Fu, Zhang, Wang, Zhou, Lin, and He}]{10.1145/3616855.3635772}
Cheng Deng, Tianhang Zhang, Zhongmou He, Qiyuan Chen, Yuanyuan Shi, Yi~Xu, Luoyi Fu, Weinan Zhang, Xinbing Wang, Chenghu Zhou, Zhouhan Lin, and Junxian He. 2024.
\newblock \href {https://doi.org/10.1145/3616855.3635772} {K2: A foundation language model for geoscience knowledge understanding and utilization}.
\newblock In \emph{Proceedings of the ACM International Conference on Web Search and Data Mining}, page 161–170.

\bibitem[{FLOCK4H(2023)}]{FlytechPythonCodes25k}
FLOCK4H. 2023.
\newblock \href {https://huggingface.co/datasets/flytech/python-codes-25k} {python-codes-25k: A python code instruction dataset}.

\bibitem[{Friel and Sanyal(2023)}]{friel2023chainpoll}
Robert Friel and Atindriyo Sanyal. 2023.
\newblock \href {https://arxiv.org/abs/2310.18344} {Chainpoll: A high efficacy method for llm hallucination detection}.
\newblock \emph{arXiv preprint arXiv:2310.18344}.

\bibitem[{Guo et~al.(2025)Guo, Yang, Zhang, Song, Wang, Zhu, Xu, Zhang, Ma, Bi, Zhang, Yu, Wu, Wu, Gou, Shao, Li, Gao, Liu, Xue, Wang, Wu, Feng, Lu, Zhao, Deng, Ruan, Dai, Chen, Ji, Li, Lin, Dai, Luo, Hao, Chen, Li, Zhang, Xu, Ding, Gao, Qu, Li, Guo, Li, Chen, Yuan, Tu, Qiu, Li, Cai, Ni, Liang, Chen, Dong, Hu, You, Gao, Guan, Huang, Yu, Wang, Zhang, Zhao, Wang, Zhang, Xu, Xia, Zhang, Zhang, Tang, Zhou, Li, Wang, Li, Tian, Huang, Zhang, Wang, Chen, Du, Ge, Zhang, Pan, Wang, Chen, Jin, Chen, Lu, Zhou, Chen, Ye, Wang, Yu, Zhou, Pan, Li, Zhou, Wu, Yun, Pei, Sun, Wang, Zeng, Liu, Liang, Gao, Yu, Zhang, Xiao, An, Liu, Wang, Chen, Nie, Cheng, Liu, Xie, Liu, Yang, Li, Su, Lin, Li, Jin, Shen, Chen, Sun, Wang, Song, Zhou, Wang, Shan, Li, Wang, Wei, Zhang, Xu, Li, Zhao, Sun, Wang, Yu, Zhang, Shi, Xiong, He, Piao, Wang, Tan, Ma, Liu, Guo, Ou, Wang, Gong, Zou, He, Xiong, Luo, You, Liu, Zhou, Zhu, Huang, Li, Zheng, Zhu, Ma, Tang, Zha, Yan, Ren, Ren, Sha, Fu, Xu, Xie, Zhang, Hao, Ma, Yan, Wu, Gu, Zhu, Liu, Li, Xie, Song,
  Pan, Huang, Xu, Zhang, and Zhang}]{guo2025deepseek}
Daya Guo, Dejian Yang, Haowei Zhang, Junxiao Song, Peiyi Wang, Qihao Zhu, Runxin Xu, Ruoyu Zhang, Shirong Ma, Xiao Bi, Xiaokang Zhang, Xingkai Yu, Yu~Wu, Z.~F. Wu, Zhibin Gou, Zhihong Shao, Zhuoshu Li, Ziyi Gao, Aixin Liu, and 175 others. 2025.
\newblock \href {http://dx.doi.org/10.1038/s41586-025-09422-z} {Deepseek-r1 incentivizes reasoning in llms through reinforcement learning}.
\newblock \emph{Nature}, 645:633–638.

\bibitem[{Guo et~al.(2024)Guo, Chen, Wang, Chang, Pei, Chawla, Wiest, and Zhang}]{guo2024large}
Taicheng Guo, Xiuying Chen, Yaqi Wang, Ruidi Chang, Shichao Pei, Nitesh~V Chawla, Olaf Wiest, and Xiangliang Zhang. 2024.
\newblock \href {https://arxiv.org/abs/2402.01680} {Large language model based multi-agents: A survey of progress and challenges}.
\newblock \emph{arXiv preprint arXiv:2402.01680}.

\bibitem[{Hilpert et~al.(2018)Hilpert, Kaldemeyer, Krien, Günther, Wingenbach, and Plessmann}]{Hilpert_2018}
S.~Hilpert, C.~Kaldemeyer, U.~Krien, S.~Günther, C.~Wingenbach, and G.~Plessmann. 2018.
\newblock \href {https://doi.org/10.1016/j.esr.2018.07.001} {The open energy modelling framework (oemof) - a new approach to facilitate open science in energy system modelling}.
\newblock \emph{Energy Strategy Reviews}, 22:16–25.

\bibitem[{Howells et~al.(2011)Howells, Rogner, Strachan, Heaps, Huntington, Kypreos, Silveira, DeCarolis, Bazillian, and Roehrl}]{article}
Mark Howells, Holger Rogner, Neil Strachan, Charles Heaps, Hillard Huntington, Socrates Kypreos, Semida Silveira, Joe DeCarolis, Morgan Bazillian, and Alexander Roehrl. 2011.
\newblock \href {https://doi.org/10.1016/j.enpol.2011.06.033} {Osemosys: The open source energy modeling system: An introduction to its ethos, structure and development}.
\newblock \emph{Energy Policy}, 39:5850--5870.

\bibitem[{Hu et~al.(2025)Hu, Kim, Ye, and Lu}]{hu2025applying}
Yi~Hu, Hyeonjin Kim, Kai Ye, and Ning Lu. 2025.
\newblock \href {https://www.sciencedirect.com/science/article/pii/S030626192402049X} {Applying fine-tuned llms for reducing data needs in load profile analysis}.
\newblock \emph{Applied Energy}, 377:124666.

\bibitem[{Jaech et~al.(2024)Jaech, Kalai, Lerer, Richardson, El-Kishky, Low, Helyar, Madry, Beutel, Carney, Iftimie, Karpenko, Passos, Neitz, Prokofiev, Wei, Tam, Bennett, Kumar, Saraiva, Vallone, Duberstein, Kondrich, Mishchenko, Applebaum, Jiang, Nair, Zoph, Ghorbani, Rossen, Sokolowsky, Barak, McGrew, Minaiev, Hao, Baker, Houghton, McKinzie, Eastman, Lugaresi, Bassin, Hudson, Li, de~Bourcy, Voss, Shen, Zhang, Koch, Orsinger, Hesse, Fischer, Chan, Roberts, Kappler, Levy, Selsam, Dohan, Farhi, Mely, Robinson, Tsipras, Li, Oprica, Freeman, Zhang, Wong, Proehl, Cheung, Mitchell, Wallace, Ritter, Mays, Wang, Such, Raso, Leoni, Tsimpourlas, Song, von Lohmann, Sulit, Salmon, Parascandolo, Chabot, Zhao, Brockman, Leclerc, Salman, Bao, Sheng, Andrin, Bagherinezhad, Ren, Lightman, Chung, Kivlichan, O'Connell, Osband, Gilaberte, Akkaya, Kostrikov, Sutskever, Kofman, Pachocki, Lennon, Wei, Harb, Twore, Feng, Yu, Weng, Tang, Yu, Candela, Palermo, Parish, Heidecke, Hallman, Rizzo, Gordon, Uesato, Ward, Huizinga, Wang,
  Chen, Xiao, Singhal, Nguyen, Cobbe, Shi, Wood, Rimbach, Gu-Lemberg, Liu, Lu, Stone, Yu, Ahmad, Yang, Liu, Maksin, Ho, Fedus, Weng, Li, McCallum, Held, Kuhn, Kondraciuk, Kaiser, Metz, Boyd, Trebacz, Joglekar, Chen, Tintor, Meyer, Jones, Kaufer, Schwarzer, Shah, Yatbaz, Guan, Xu, Yan, Glaese, Chen, Lampe, Malek, Wang, Fradin, McClay, Pavlov, Wang, Wang, Murati, Bavarian, Rohaninejad, McAleese, Chowdhury, Chowdhury, Ryder, Tezak, Brown, Nachum, Boiko, Murk, Watkins, Chao, Ashbourne, Izmailov, Zhokhov, Dias, Arora, Lin, Lopes, Gaon, Miyara, Leike, Hwang, Garg, Brown, James, Shu, Cheu, Greene, Jain, Altman, Toizer, Toyer, Miserendino, Agarwal, Hernandez, Baker, McKinney, Yan, Zhao, Hu, Santurkar, Chaudhuri, Zhang, Fu, Papay, Lin, Balaji, Sanjeev, Sidor, Broda, Clark, Wang, Gordon, Sanders, Patwardhan, Sottiaux, Degry, Dimson, Zheng, Garipov, Stasi, Bansal, Creech, Peterson, Eloundou, Qi, Kosaraju, Monaco, Pong, Fomenko, Zheng, Zhou, McCabe, Zaremba, Dubois, Lu, Chen, Cha, Bai, He, Zhang, Wang, Shao, and
  Li}]{jaech2024openai}
Aaron Jaech, Adam Kalai, Adam Lerer, Adam Richardson, Ahmed El-Kishky, Aiden Low, Alec Helyar, Aleksander Madry, Alex Beutel, Alex Carney, Alex Iftimie, Alex Karpenko, Alex~Tachard Passos, Alexander Neitz, Alexander Prokofiev, Alexander Wei, Allison Tam, Ally Bennett, Ananya Kumar, and 242 others. 2024.
\newblock \href {https://arxiv.org/abs/2412.16720} {Openai o1 system card}.
\newblock \emph{arXiv preprint arXiv:2412.16720}.

\bibitem[{Jiang et~al.(2024)Jiang, Ma, Zhang, and Chen}]{jiang2024eplus}
Gang Jiang, Zhihao Ma, Liang Zhang, and Jianli Chen. 2024.
\newblock \href {https://www.sciencedirect.com/science/article/pii/S0306261924008146} {Eplus-llm: A large language model-based computing platform for automated building energy modeling}.
\newblock \emph{Applied Energy}, 367:123431.

\bibitem[{Jiang et~al.(2025{\natexlab{a}})Jiang, Cheng, Moreira, Zhu, Sun, Ren, He, Dai, and Hua}]{jiang2025ucdr}
Haoyu Jiang, Zhi-Qi Cheng, Gabriel Moreira, Jiawen Zhu, Jingdong Sun, Bukun Ren, Jun-Yan He, Qi~Dai, and Xian-Sheng Hua. 2025{\natexlab{a}}.
\newblock \href {https://ieeexplore.ieee.org/abstract/document/10943400} {Ucdr-adapter: Exploring adaptation of pre-trained vision-language models for universal cross-domain retrieval}.
\newblock In \emph{Proceedings of the IEEE/CVF Winter Conference on Applications of Computer Vision}, pages 5429--5438.

\bibitem[{Jiang et~al.(2025{\natexlab{b}})Jiang, Qu, Zhu, Zeng, Lin, and Zhong}]{jiang2025hyperloadcrossmodalityenhancedlarge}
Haoyu Jiang, Boan Qu, Junjie Zhu, Fanjie Zeng, Xiaojie Lin, and Wei Zhong. 2025{\natexlab{b}}.
\newblock \href {https://arxiv.org/abs/2512.19114} {Hyperload: A cross-modality enhanced large language model-based framework for green data center cooling load prediction}.
\newblock \emph{Preprint}, arXiv:2512.19114.

\bibitem[{Jin et~al.(2023)Jin, Wang, Ma, Chu, Zhang, Shi, Chen, Liang, Li, Pan, and Wen}]{jin2023time}
Ming Jin, Shiyu Wang, Lintao Ma, Zhixuan Chu, James~Y. Zhang, Xiaoming Shi, Pin-Yu Chen, Yuxuan Liang, Yuan-Fang Li, Shirui Pan, and Qingsong Wen. 2023.
\newblock \href {https://arxiv.org/abs/2310.01728} {Time-llm: Time series forecasting by reprogramming large language models}.

\bibitem[{Lerede et~al.(2024)Lerede, Cosmo, and Savoldi}]{Lerede2024TEMOAEuropeAO}
Daniele Lerede, Valeria~Di Cosmo, and Laura Savoldi. 2024.
\newblock \href {https://www.sciencedirect.com/science/article/pii/S0360544224026240?via%3Dihub} {Temoa-europe: an open-source and open-data energy system optimization model for the analysis of the european energy mix}.
\newblock \emph{Energy}, 308:132850.

\bibitem[{Li et~al.(2023)Li, Hammoud, Itani, Khizbullin, and Ghanem}]{li2023camel}
Guohao Li, Hasan Hammoud, Hani Itani, Dmitrii Khizbullin, and Bernard Ghanem. 2023.
\newblock \href {https://proceedings.neurips.cc/paper_files/paper/2023/file/a3621ee907def47c1b952ade25c67698-Paper-Conference.pdf} {Camel: Communicative agents for "mind" exploration of large language model society}.
\newblock In \emph{Proceedings of the Advances in Neural Information Processing Systems}, volume~36, pages 51991--52008.

\bibitem[{Li(2023)}]{ZihaoLi2023IEAEnergyDataset}
Zihao Li. 2023.
\newblock \href {https://huggingface.co/datasets/Zihao-Li/IEA_Energy_Dataset} {Iea energy dataset}.

\bibitem[{Liao et~al.(2025)Liao, Wang, Yang, Yang, Fang, Rehtanz, and Port{\'e}-Agel}]{liao2025timegpt}
Wenlong Liao, Shouxiang Wang, Dechang Yang, Zhe Yang, Jiannong Fang, Christian Rehtanz, and Fernando Port{\'e}-Agel. 2025.
\newblock \href {https://www.sciencedirect.com/science/article/pii/S0306261924023572} {Timegpt in load forecasting: A large time series model perspective}.
\newblock \emph{Applied Energy}, 379:124973.

\bibitem[{Lin et~al.(2025{\natexlab{a}})Lin, Zhang, Li, Yuan, Yu, Li, He, Jiang, Li, Song, Tang, Xiao, Lin, Zhuang, and Ooi}]{Lin2025HealthGPTAM}
Tianwei Lin, Wenqiao Zhang, Sijing Li, Yuqian Yuan, Binhe Yu, Haoyuan Li, Wanggui He, Hao Jiang, Mengze Li, Xiaohui Song, Siliang Tang, Jun Xiao, Hui Lin, Yueting Zhuang, and Bengchin Ooi. 2025{\natexlab{a}}.
\newblock \href {https://arxiv.org/abs/2502.09838} {Healthgpt: A medical large vision-language model for unifying comprehension and generation via heterogeneous knowledge adaptation}.
\newblock In \emph{Proceedings of the International Conference on Machine Learning}.

\bibitem[{Lin et~al.(2025{\natexlab{b}})Lin, Luo, Du-Ikonen, Lin, Mao, Jiang, Wang, Yuan, Zhong, and Yu}]{lin2025generative}
Xiaojie Lin, Zheng Luo, Liuliu Du-Ikonen, Xueru Lin, Yihui Mao, Haoyu Jiang, Shuai Wang, Chongshuo Yuan, Wei Zhong, and Zitao Yu. 2025{\natexlab{b}}.
\newblock \href {https://www.sciencedirect.com/science/article/pii/S2666546825001429} {Generative artificial intelligence: Pioneering a new paradigm for research and education in smart energy systems}.
\newblock \emph{Energy and AI}, 22:100610.

\bibitem[{Lund et~al.(2017)Lund, {\O}stergaard, Connolly, and Mathiesen}]{lund2017smart}
Henrik Lund, Poul~Alberg {\O}stergaard, David Connolly, and Brian~Vad Mathiesen. 2017.
\newblock \href {https://www.sciencedirect.com/science/article/pii/S0360544217308812} {Smart energy and smart energy systems}.
\newblock \emph{Energy}, 137:556--565.

\bibitem[{Majidi et~al.(2025)Majidi, Hayati, Breyer, Mohammadi-ivatloo, Honkapuro, Karjunen, Laaksonen, and Sihvonen}]{RePEc:eee:rensus:v:212:y:2025:i:c:s1364032125000401}
Hassan Majidi, Mohammad~Mohsen Hayati, Christian Breyer, Behnam Mohammadi-ivatloo, Samuli Honkapuro, Hannu Karjunen, Petteri Laaksonen, and Ville Sihvonen. 2025.
\newblock \href {https://www.sciencedirect.com/science/article/pii/S1364032125000401} {{Overview of energy modeling requirements and tools for future smart energy systems}}.
\newblock \emph{Renewable and Sustainable Energy Reviews}, 212:115367.

\bibitem[{Mao et~al.(2023)Mao, Cai, Xia, Wu, Wang, Wang, Ge, and Wei}]{mao2023alympics}
Shaoguang Mao, Yuzhe Cai, Yan Xia, Wenshan Wu, Xun Wang, Fengyi Wang, Tao Ge, and Furu Wei. 2023.
\newblock \href {https://aclanthology.org/2025.coling-main.193/} {Alympics: Llm agents meet game theory}.
\newblock In \emph{Proceedings of the International Conference on Computational Linguistics}, pages 2845--2866.

\bibitem[{Mishra et~al.(2022)Mishra, Khashabi, Baral, and Hajishirzi}]{naturalinstructions}
Swaroop Mishra, Daniel Khashabi, Chitta Baral, and Hannaneh Hajishirzi. 2022.
\newblock \href {https://aclanthology.org/2022.acl-long.244/} {Cross-task generalization via natural language crowdsourcing instructions}.
\newblock In \emph{Proceedings of the International Conference on Computational Linguistics}.

\bibitem[{Muennighoff(2022)}]{MuennighoffNaturalInstructions}
Niklas Muennighoff. 2022.
\newblock \href {https://huggingface.co/datasets/Muennighoff/natural-instructions} {natural-instructions: Preprocessed version of super-natural-instructions}.

\bibitem[{Nam et~al.(2024)Nam, Macvean, Hellendoorn, Vasilescu, and Myers}]{nam2024using}
Daye Nam, Andrew Macvean, Vincent Hellendoorn, Bogdan Vasilescu, and Brad Myers. 2024.
\newblock \href {https://dl.acm.org/doi/abs/10.1145/3597503.3639187} {Using an llm to help with code understanding}.
\newblock In \emph{Proceedings of the IEEE/ACM International Conference on Software Engineering}, pages 1--13.

\bibitem[{Ni and Buehler(2024)}]{ni2024mechagents}
Bo~Ni and Markus~J Buehler. 2024.
\newblock \href {https://www.sciencedirect.com/science/article/pii/S2352431624000117} {Mechagents: Large language model multi-agent collaborations can solve mechanics problems, generate new data, and integrate knowledge}.
\newblock \emph{Extreme Mechanics Letters}, 67:102131.

\bibitem[{Park et~al.(2023)Park, O'Brien, Cai, Morris, Liang, and Bernstein}]{park2023generative}
Joon~Sung Park, Joseph O'Brien, Carrie~Jun Cai, Meredith~Ringel Morris, Percy Liang, and Michael~S Bernstein. 2023.
\newblock \href {https://dl.acm.org/doi/abs/10.1145/3586183.3606763} {Generative agents: Interactive simulacra of human behavior}.
\newblock In \emph{Proceedings of the ACM Symposium on User Interface Software and Technology}, pages 1--22.

\bibitem[{Paruchuri(2025)}]{paruchuri2025marker}
Vikram Paruchuri. 2025.
\newblock \href {https://github.com/VikParuchuri/marker} {Marker: Convert pdf to markdown + json quickly with high accuracy}.

\bibitem[{Qin et~al.(2023)Qin, Liang, Ye, Zhu, Yan, Lu, Lin, Cong, Tang, Qian, Zhao, Tian, Xie, Zhou, Gerstein, Li, Liu, and Sun}]{qin2023toolllm}
Yujia Qin, Shihao Liang, Yining Ye, Kunlun Zhu, Lan Yan, Yaxi Lu, Yankai Lin, Xin Cong, Xiangru Tang, Bill Qian, Sihan Zhao, Runchu Tian, Ruobing Xie, Jie Zhou, Mark Gerstein, Dahai Li, Zhiyuan Liu, and Maosong Sun. 2023.
\newblock \href {https://arxiv.org/abs/2307.16789} {Toolllm: Facilitating large language models to master 16000+ real-world apis}.
\newblock In \emph{Proceedings of the International Conference on Learning Representations}.

\bibitem[{R1(2025)}]{OpenR1Math220k}
Open R1. 2025.
\newblock \href {https://huggingface.co/datasets/open-r1/OpenR1-Math-220k} {Openr1-math-220k: A large-scale dataset for mathematical reasoning}.

\bibitem[{Taori et~al.(2023{\natexlab{a}})Taori, Gulrajani, Zhang, Dubois, Li, Guestrin, Liang, and Hashimoto}]{taori2023alpaca}
Rohan Taori, Ishaan Gulrajani, Tianyi Zhang, Yann Dubois, Xuechen Li, Carlos Guestrin, Percy Liang, and Tatsunori~B Hashimoto. 2023{\natexlab{a}}.
\newblock \href {https://crfm.stanford.edu/2023/03/13/alpaca.html} {Alpaca: A strong, replicable instruction-following model}.

\bibitem[{Taori et~al.(2023{\natexlab{b}})Taori, Gulrajani, Zhang, Dubois, Li, Guestrin, Liang, and Hashimoto}]{alpaca}
Rohan Taori, Ishaan Gulrajani, Tianyi Zhang, Yann Dubois, Xuechen Li, Carlos Guestrin, Percy Liang, and Tatsunori~B. Hashimoto. 2023{\natexlab{b}}.
\newblock \href {https://github.com/tatsu-lab/stanford_alpac} {Stanford alpaca: An instruction-following llama model}.

\bibitem[{Thellufsen et~al.(2020)Thellufsen, Lund, Sorkn{\ae}s, {\O}stergaard, Chang, Drysdale, Nielsen, Dj{\o}rup, and Sperling}]{thellufsen2020smart}
Jakob~Zinck Thellufsen, Henrik Lund, P~Sorkn{\ae}s, PA~{\O}stergaard, M~Chang, D~Drysdale, Steen Nielsen, SR~Dj{\o}rup, and K~Sperling. 2020.
\newblock \href {https://www.sciencedirect.com/science/article/pii/S1364032120302136} {Smart energy cities in a 100\% renewable energy context}.
\newblock \emph{Renewable and Sustainable Energy Reviews}, 129:109922.

\bibitem[{Tian et~al.(2024)Tian, Gan, Song, Zhang, and Zhang}]{tian-etal-2024-chimed}
Yuanhe Tian, Ruyi Gan, Yan Song, Jiaxing Zhang, and Yongdong Zhang. 2024.
\newblock \href {https://aclanthology.org/2024.acl-long.386.pdf} {{C}hi{M}ed-{GPT}: A {C}hinese medical large language model with full training regime and better alignment to human preferences}.
\newblock In \emph{Proceedings of the Annual Meeting of the Association for Computational Linguistics (Volume 1: Long Papers)}, pages 7156--7173.

\bibitem[{Tirumala et~al.(2023)Tirumala, Simig, Aghajanyan, and Morcos}]{tirumala2023d4}
Kushal Tirumala, Daniel Simig, Armen Aghajanyan, and Ari~S Morcos. 2023.
\newblock \href {https://proceedings.neurips.cc/paper_files/paper/2023/hash/a8f8cbd7f7a5fb2c837e578c75e5b615-Abstract-Datasets_and_Benchmarks.html} {D4: improving llm pretraining via document de-duplication and diversification}.
\newblock In \emph{Proceedings of the International Conference on Neural Information Processing Systems}, pages 53983--53995.

\bibitem[{Wang et~al.(2025{\natexlab{a}})Wang, Qi, Wang, Sun, Zhuang, Wu, Zhang, and Liao}]{wang2025chattime}
Chengsen Wang, Qi~Qi, Jingyu Wang, Haifeng Sun, Zirui Zhuang, Jinming Wu, Lei Zhang, and Jianxin Liao. 2025{\natexlab{a}}.
\newblock \href {https://ojs.aaai.org/index.php/AAAI/article/view/33384} {Chattime: A unified multimodal time series foundation model bridging numerical and textual data}.
\newblock In \emph{Proceedings of the AAAI Conference on Artificial Intelligence}, volume~39, pages 12694--12702.

\bibitem[{Wang et~al.(2025{\natexlab{b}})Wang, Zhou, Liang, Yu, and Jing}]{wang2025climate}
Meng Wang, Jingfeng Zhou, Yujing Liang, Hang Yu, and Rui Jing. 2025{\natexlab{b}}.
\newblock \href {https://www.sciencedirect.com/science/article/pii/S2210670725002082} {Climate change impacts on city-scale building energy performance based on gis-informed urban building energy modelling}.
\newblock \emph{Sustainable Cities and Society}, 125:106331.

\bibitem[{Wang et~al.(2023)Wang, Kordi, Mishra, Liu, Smith, Khashabi, and Hajishirzi}]{wang2023self}
Yizhong Wang, Yeganeh Kordi, Swaroop Mishra, Alisa Liu, Noah~A. Smith, Daniel Khashabi, and Hannaneh Hajishirzi. 2023.
\newblock \href {https://aclanthology.org/2023.acl-long.754/} {Self-instruct: Aligning language models with self-generated instructions}.
\newblock In \emph{Proceedings of the Annual Meeting of the Association for Computational Linguistics (Volume 1: Long Papers)}.

\bibitem[{Wang et~al.(2022{\natexlab{a}})Wang, Mishra, Alipoormolabashi, Kordi, Mirzaei, Arunkumar, Ashok, Dhanasekaran, Naik, Stap, Pathak, Karamanolakis, Lai, Purohit, Mondal, Anderson, Kuznia, Doshi, Patel, Pal, Moradshahi, Parmar, Purohit, Varshney, Kaza, Verma, Puri, Karia, Sampat, Doshi, Mishra, Reddy, Patro, Dixit, Shen, Baral, Choi, Smith, Hajishirzi, and Khashabi}]{supernaturalinstructions}
Yizhong Wang, Swaroop Mishra, Pegah Alipoormolabashi, Yeganeh Kordi, Amirreza Mirzaei, Anjana Arunkumar, Arjun Ashok, Arut~Selvan Dhanasekaran, Atharva Naik, David Stap, Eshaan Pathak, Giannis Karamanolakis, Haizhi~Gary Lai, Ishan Purohit, Ishani Mondal, Jacob Anderson, Kirby Kuznia, Krima Doshi, Maitreya Patel, and 21 others. 2022{\natexlab{a}}.
\newblock \href {https://par.nsf.gov/servlets/purl/10462307} {Super-naturalinstructions:generalization via declarative instructions on 1600+ tasks}.
\newblock In \emph{Proceedings of the Conference on Empirical Methods in Natural Language Processing}.

\bibitem[{Wang et~al.(2022{\natexlab{b}})Wang, Zhou, Wang, Lin, Shah, and Lim}]{wang2022few}
Ze~Wang, Yipin Zhou, Rui Wang, Tsung-Yu Lin, Ashish Shah, and Ser~Nam Lim. 2022{\natexlab{b}}.
\newblock \href {https://proceedings.neurips.cc/paper_files/paper/2022/hash/1fe6f635fe265292aba3987b5123ae3d-Abstract-Conference.html} {Few-shot fast-adaptive anomaly detection}.
\newblock \emph{Proceedings of the Advances in Neural Information Processing Systems}, 35:4957--4970.

\bibitem[{Wu and Ling(2024)}]{wu2024stellm}
Tangjie Wu and Qiang Ling. 2024.
\newblock \href {https://www.sciencedirect.com/science/article/pii/S030626192401417X} {Stellm: Spatio-temporal enhanced pre-trained large language model for wind speed forecasting}.
\newblock \emph{Applied Energy}, 375:124034.

\bibitem[{Xu et~al.(2023)Xu, Yu, Fang, Wang, and Wu}]{xu2023language}
Zelai Xu, Chao Yu, Fei Fang, Yu~Wang, and Yi~Wu. 2023.
\newblock \href {https://arxiv.org/abs/2310.18940} {Language agents with reinforcement learning for strategic play in the werewolf game}.
\newblock In \emph{Proceedings of the International Conference on Machine Learning}.

\bibitem[{Yang et~al.(2024)Yang, Yang, Zhang, Hui, Zheng, Yu, Li, Liu, Huang, Wei, Lin, Yang, Tu, Zhang, Yang, Yang, Zhou, Lin, Dang, Lu, Bao, Yang, Yu, Li, Xue, Zhang, Zhu, Men, Lin, Li, Tang, Xia, Ren, Ren, Fan, Su, Zhang, Wan, Liu, Cui, Zhang, and Qiu}]{yang2024qwen2}
An~Yang, Baosong Yang, Beichen Zhang, Binyuan Hui, Bo~Zheng, Bowen Yu, Chengyuan Li, Dayiheng Liu, Fei Huang, Haoran Wei, Huan Lin, Jian Yang, Jianhong Tu, Jianwei Zhang, Jianxin Yang, Jiaxi Yang, Jingren Zhou, Junyang Lin, Kai Dang, and 23 others. 2024.
\newblock \href {https://arxiv.org/abs/2412.15115} {Qwen2. 5 technical report}.
\newblock \emph{arXiv preprint arXiv:2412.15115}.

\bibitem[{Zhang et~al.(2024)Zhang, Liu, Tan, Chen, Yan, Yan, Li, Huang, Yue, Ouyang, Zhou, Zhang, Su, Zhong, and Li}]{zhang2024chemllmchemicallargelanguage}
Di~Zhang, Wei Liu, Qian Tan, Jingdan Chen, Hang Yan, Yuliang Yan, Jiatong Li, Weiran Huang, Xiangyu Yue, Wanli Ouyang, Dongzhan Zhou, Shufei Zhang, Mao Su, Han-Sen Zhong, and Yuqiang Li. 2024.
\newblock \href {https://arxiv.org/abs/2402.06852} {Chemllm: A chemical large language model}.
\newblock \emph{arXiv preprint arXiv:2402.06852}.

\bibitem[{Zhang et~al.(2025)Zhang, Zhang, Lu, and Zhao}]{zhang2025domain}
Jian Zhang, Chaobo Zhang, Jie Lu, and Yang Zhao. 2025.
\newblock \href {https://www.sciencedirect.com/science/article/pii/S0306261924017616} {Domain-specific large language models for fault diagnosis of heating, ventilation, and air conditioning systems by labeled-data-supervised fine-tuning}.
\newblock \emph{Applied Energy}, 377:124378.

\bibitem[{Zhang et~al.(2023)Zhang, Dong, Li, Zhang, Sun, Wang, Li, Hu, Zhang, and Wu}]{zhang2023instruction}
Shengyu Zhang, Linfeng Dong, Xiaoya Li, Sen Zhang, Xiaofei Sun, Shuhe Wang, Jiwei Li, Runyi Hu, Tianwei Zhang, and Fei Wu. 2023.
\newblock \href {https://dl.acm.org/doi/full/10.1145/3777411} {Instruction tuning for large language models: A survey}.
\newblock \emph{ACM Computing Surveys}.

\bibitem[{Zheng et~al.(2025)Zheng, Koh, Ju, Nguyen, May, Webb, and Pan}]{zheng2025large}
Yizhen Zheng, Huan~Yee Koh, Jiaxin Ju, Anh~TN Nguyen, Lauren~T May, Geoffrey~I Webb, and Shirui Pan. 2025.
\newblock \href {https://www.nature.com/articles/s42256-025-00994-z} {Large language models for scientific discovery in molecular property prediction}.
\newblock \emph{Nature Machine Intelligence}, pages 1--11.

\end{thebibliography}

\end{document}